\documentclass[10pt,twocolumn,letterpaper]{article}

\usepackage{iccv}
\usepackage{times}
\usepackage{epsfig}
\usepackage{graphicx}
\usepackage{amsmath}
\usepackage{amssymb}

\usepackage{booktabs}       
\usepackage{multirow}
\usepackage{tabularx}
\usepackage{hhline}
\usepackage{paralist}
\usepackage{amsmath}
\usepackage{amsfonts} 

\usepackage{epsfig}
\usepackage{amsmath}
\usepackage{amssymb}
\usepackage{adjustbox}
\usepackage{amsthm}  
\usepackage{wasysym}
\usepackage[ruled,vlined,algo2e]{algorithm2e}

\usepackage{color}
\usepackage{dsfont}	
\usepackage{footnote}
\usepackage{epstopdf}
\usepackage{enumitem}
\usepackage{subfigure}
\usepackage{caption}
\usepackage{comment}
\usepackage{multirow}
\usepackage{algorithm}
\usepackage{algorithmic}
\usepackage[accsupp]{axessibility}

\def\eg{\emph{e.g., }}
\def\ie{\emph{i.e., }}

\def\vs{\emph{vs. }}
\def\wrt{\emph{w.r.t. }}
\def\etal{\emph{et al. }}

\usepackage{pifont}

\usepackage{mathtools}

\makeatletter
\newcommand*{\rom}[1]{\expandafter\@slowromancap\romannumeral #1@}
\makeatother
\makeatletter
\newcommand\footnoteref[1]{\protected@xdef\@thefnmark{\ref{#1}}\@footnotemark}
\makeatother
\newcommand{\bfsection}[1]{\vspace*{0.1cm}\noindent\textbf{#1.}}

\usepackage[pagebackref=true,breaklinks=true,letterpaper=true,colorlinks,bookmarks=false]{hyperref}

\iccvfinalcopy 

\def\iccvPaperID{5524} 

\ificcvfinal\fi

\begin{document}

\title{Self-supervised Geometric Features Discovery via Interpretable Attention \\for Vehicle Re-Identification and Beyond}

\author{Ming Li\thanks{This work was done when the author visited VISLab at WPI \url{https://zhang-vislab.github.io}} \hspace{2cm} Xinming Huang \hspace{2cm} Ziming Zhang\\
Worcester Polytechnic Institute\\
100 Institute Rd, Worcester, MA, USA\\
{\tt\small ming.li@u.nus.edu, \{xhuang, zzhang15\}@wpi.edu}
}

\maketitle
\ificcvfinal\thispagestyle{empty}\fi

\begin{abstract}
   To learn distinguishable patterns, most of recent works in vehicle re-identification (ReID) struggled to redevelop official benchmarks to provide various supervisions, which requires prohibitive human labors. In this paper, we seek to achieve the similar goal but do not involve more human efforts. To this end, we introduce a novel framework, which successfully encodes both geometric local features and global representations to distinguish vehicle instances, optimized only by the supervision from official ID labels. Specifically, given our insight that objects in ReID share similar geometric characteristics, we propose to borrow self-supervised representation learning to facilitate geometric features discovery. To condense these features, we introduce an interpretable attention module, with the core of local maxima aggregation instead of fully automatic learning, whose mechanism is completely understandable and whose response map is physically reasonable. To the best of our knowledge, we are the first that perform self-supervised learning to discover geometric features. We conduct comprehensive experiments on three most popular datasets for vehicle ReID, \ie VeRi-776, CityFlow-ReID, and VehicleID. We report our state-of-the-art (SOTA) performances and promising visualization results. We also show the excellent scalability of our approach on other ReID related tasks, \ie person ReID and multi-target multi-camera (MTMC) vehicle tracking. The code is available at \url{https://github.com/ming1993li/Self-supervised-Geometric}.
\end{abstract} 

\section{Introduction}
Vehicle ReID is a fundamental but challenging problem in video surveillance due to subtle discrepancy among vehicles from identical make and large variation across viewpoints of the same instance. The success of recent works suggests that the key to solving this problem is to incorporate
\begin{figure}[t]
\centering
\includegraphics[width=0.99\columnwidth]{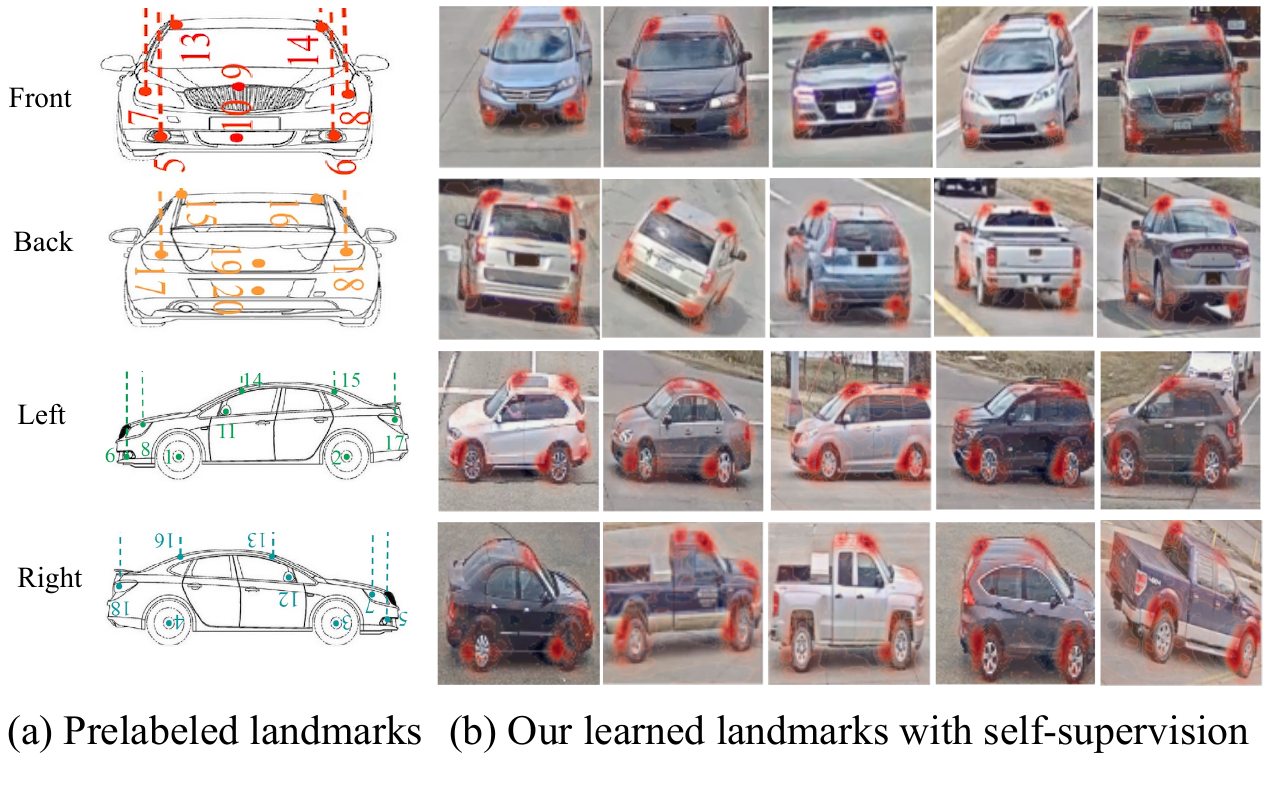}
\includegraphics[width=0.99\columnwidth]{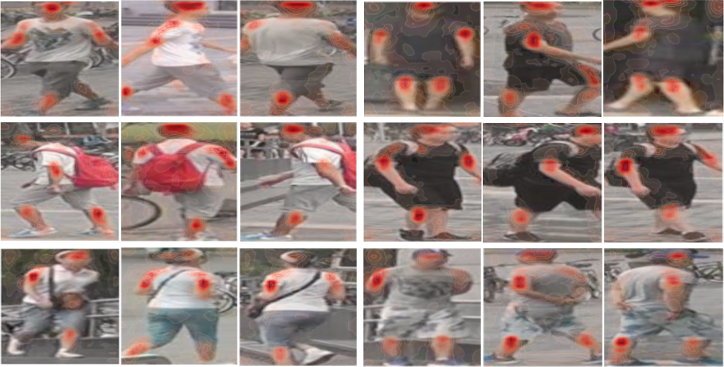}
\caption{ {\bf (Top)} In the literature, a labor-intensive fine-grained labels annotation is often required to capture local discriminative features, such as (a) labelling landmarks in \cite{key_points} to learn orientation invariant features. In contrast, we manage to discover such geometric features (denoted as red in (b)) in a self-supervised way. {\bf (Bottom)} As for its generalization ability, our approach can also consistently locate critical parts of a deformable human body, \eg head, upper arms, and knees, with no using corresponding ground truths. Best viewed in color.}
\vspace{-5mm}
\label{fig-fig1}
\end{figure}
explicit mechanisms to discover and concentrate on informative vehicle parts (\eg wheels, manufacturer logos) for discriminative feature extraction in addition to capturing robust features from a holistic image. They all sought, however, to edit original data to provide supplementary supervisions, \eg{view segmentation \cite{parsing2020}, keypoints \cite{dual_path, key_points}, vehicle orientation \cite{dual_path, key_points, vanet, chen2020orientation} or key parts \cite{part_regularized}}, for training their deep classifiers. Although these methods perform satisfactorily, their annotation processes inevitably involve intensive human efforts, which significantly limits the applicability of such approaches. For example, when deployed in a new scenario, \cite{part_regularized} demands the informative parts have to be manually localized to optimize their YOLO detector. Afterwards, they are able to embed local patterns from detected regions-of-interest to assist ReID. So it is desirable to develop approaches which are capable of concentrating on informative details of a vehicle body but do not require corresponding ground truths.

On the other hand, while their power is demonstrated by various computer vision tasks \cite{dual_attention, Chen_2019_ICCV, dual_path, abdnet}, existing attention mechanisms, like channel attention \cite{senet}, spatial attention \cite{cbam}, and self-attention \cite{attentions}, are all pretty sophisticated and obscure. That is to say, their architectures are difficult to explain and attention maps are learned all by themselves. In self-attention \cite{attentions}, for example, high-dimensional embeddings $Q$, $K$, $V$ are first projected from an input by convolutional or linear operations and then entry-wise correlation (attention) is obtained through matrix multiplication between $Q$, $K$. $V$ is weighted by the resulting correlation matrix as the attentional output. Although the workflow seems to make sense, the underlying principle why it works is still a black-box like other deep networks. Additionally, their learned attentions usually spread over a holistic object without specific concerns. Otherwise, an interpretable attention module, whose design should be easy to understand, can reveal what is critical for recognition and help to guide further improvement. 

In light of the above observations, we propose a novel framework that 
can successfully learn discriminative geometric features, under the assistance of self-supervised learning and a simple but interpretable attention, in addition to global representations for vehicle ReID. In specific, self-supervised learning is performed to optimize an encoder network, which is shared to condense low-level vehicle representations, under the supervision of automatically generated ground truths. The encoded vehicle representations are fed into the introduced interpretable attention mechanism to acquire an attention map. By weighting it on another low-level vehicle representations, we obtain the regions-of-interest emphasized features for vehicle ReID. This is the complete version of \cite{li2021self} including the code link. 

In summary, our key contributions in this work are:
\begin{itemize}[nosep, leftmargin=*]
\item We are the first to successfully learn informative geometric features for vehicle ReID without supervisions from fine-grained annotations. 
\item An interpretable attention module, whose design is easy to explain and whose concentrations are physically important locations, is introduced to highlight the automatic regions-of-interest.
\item We report the SOTA performances of our proposed approach on widely used vehicle ReID benchmarks, \ie VeRi-776 \cite{veri}, CityFlow-ReID \cite{cityflow}, and VehicleID \cite{vehicleid}, compared with all existing works including those involving more supervisions from manual annotations. We also visualize the reliable and consistent geometric features learned by our framework.
\item The excellent scalability of the proposal is demonstrated by our directly transferring experiments on person ReID and MTMC vehicle tracking.
\end{itemize}

\section{Related works}
\bfsection{Vehicle ReID}
Most of existing works in this field struggled to explore extra supervisions in addition to identity labels to guide ReID. These works can be grouped into three mainstreams as follows: 
(1) exploiting vehicle attributes (\eg{color and model}) \cite{two_level, yan2017exploiting, liu2017provid, veri, liu2018ram, Zhou_2018_CVPR} or temporal information in data \cite{key_points, shen2017learning} to regularize representation learning; (2) editing official datasets to provide more fine-grained annotations, like critical part locations \cite{part_regularized}, view segmentation \cite{parsing2020}, keypoints or vehicle body orientation \cite{dual_path, key_points, vanet, chen2020orientation}, to supervise local feature discovery; (3) assembling multiple datasets together \cite{vehiclenet} or synthesizing more vehicle images \cite{lou2019embedding, pose_multi, wu2018joint} to train more powerful networks. Additionally, there are a couple of works aiming to enhance representation learning from the perspective of metric learning \cite{deep_meta, vanet, bai2018group, zhang2017improving}. In contrast, our work manages to capture discriminative local patterns without corresponding supervision. Furthermore, unlike recent well-performing works which relied on another auxiliary pretrained network to indicate informative parts \cite{parsing2020, part_regularized, chen2020orientation, vanet}, our framework is elegant and end-to-end trainable.

\bfsection{Visual attention}
Various attention architectures have been proposed in computer vision community, \eg{self-attention \cite{attentions, dual_attention}, channel-wise attention \cite{senet}, and spatial-wise attention \cite{cbam}}, which also spread into ReID field \cite{abdnet, Chen_2019_ICCV, zhou2019discriminative, dual_path, li2021exploiting}. For instance, \cite{Chen_2019_ICCV} and \cite{zhou2019discriminative} proposed using attention gains and multi-level foreground consistency to regularize ReID feature extraction, respectively. All these attention networks are pretty complicated and computational costly, especially hard to explain, which limits their generalization and future improvement. The attentive branch in \cite{abdnet}, for example, incorporated Channel Attention Module (CAM) and Position Attention Module (PAM) in parallel. The reason why the latter employed stacked convolutional layers to perform $Q$, $K$, $V$ projection but the former just utilized identity layer (copy) instead is unknown. In this case, we have no idea to improve it further, \eg{making the positional attention focus on more distinguishable parts rather than a large general area of human body in \cite{abdnet}}. Differently, our attention is composed of only a couple of learnable operations and each step is reasonable and easy to explain.

\begin{figure}[t]
\centering
\includegraphics[width=0.99\columnwidth]{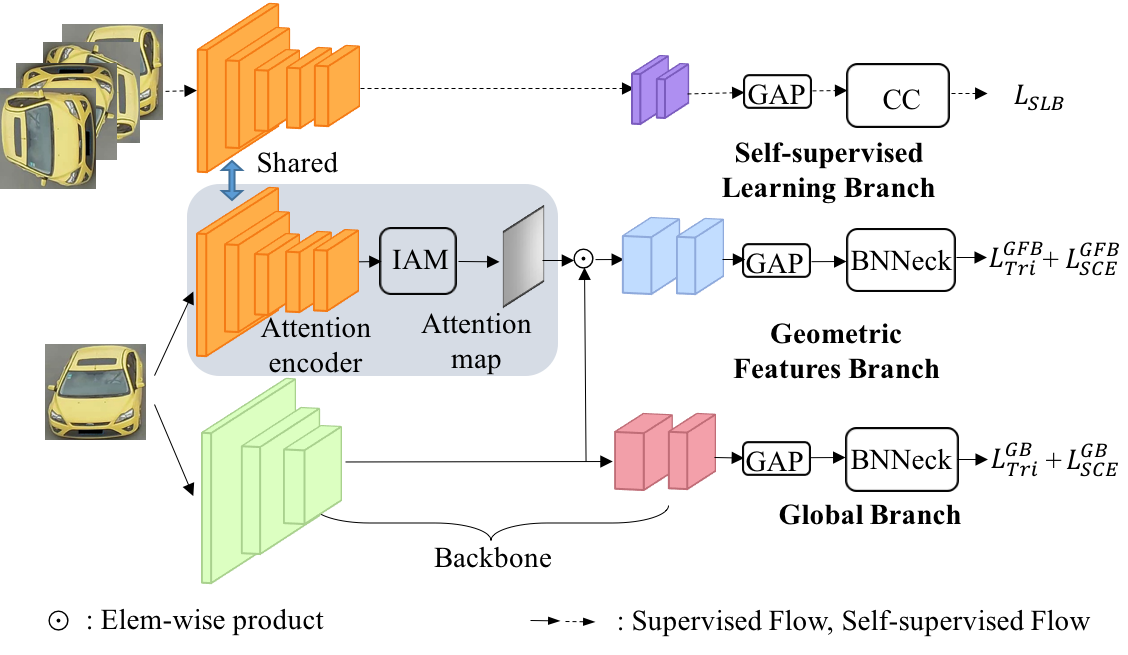} 
\caption{Overview of our framework: Self-supervised Geometric Features Discovery via Interpretable Attention, which consists of Global Branch (GB), Self-supervised Learning Branch (SLB), and Geometric Features Branch (GFB). Some key components are Interpretable Attention Module (IAM), Batch Normalization Neck (BNNeck) \cite{bag_of_tricks}, Cosine Classifier (CC) \cite{boostfewshoot}, Global Average Pooling (GAP), hard mining Triplet loss (Tri) \cite{triplet_loss}, and Smoothed Cross Entropy loss (SCE) \cite{smoothed_ce1}.}
\vspace{-6mm}
\label{fig-overview}
\end{figure}

\bfsection{Self-supervised learning}
The success of self-supervised learning hinges on devising an appropriate pretext task to supervise model optimization. In this literature, a variety of visual tasks have been constructed, for instance, image completion \cite{global_local}, colorization \cite{zhang2017split}, patch position prediction \cite{doersch2015unsupervised, Kolesnikov_2019_CVPR}, patch order prediction \cite{kim2019self, Kolesnikov_2019_CVPR} and rotation recognition \cite{Zhai_2019_ICCV, feng2019self}. Besides, recently contrastive learning of multi-view images \cite{simclr, moco} has demonstrated its efficacy. Furthermore, these pretext tasks can also be borrowed as an auxiliary task to strengthen a targeted one \cite{boostfewshoot, self_gan}. Ours is significantly different from these works. We conduct self-supervised learning to facilitate geometric features discovery, which has not been explored by other works yet.

We are aware that recently Khorramshahi \etal \cite{khorramshahi2020devil} proposed their Self-supervised Attention for Vehicle Re-identification (SAVER) framework to pay attention to details of a vehicle body. Although our title shares terms ``self-supervised'' and ``attention'' with theirs, our approach is completely different from theirs. In principle, SAVER took the residual of removing the reconstructed image by Variational Auto-Encoder (VAE) \cite{kingma2013auto} from an input as discriminative details for feature extraction, which they called ``self-supervised attention''. However, we actually propose a novel approach of borrowing self-supervised learning to regularize our interpretable attention learning. Besides, our proposal obviously differs from SAVER on these aspects, at least:
\begin{itemize}[nosep,leftmargin=*]
\item Ours can robustly and consistently locate geometric features with physical interpretation across vehicle instances and viewpoints.
\item Our deep framework is more concise with no need of extra offline pretraining (like VAE in SAVER) or customized image pre-processing, \ie removing background noise from all images using object detector.
\item Our results are significantly better. For instance, even on the most challenging testing scenario of VehicleID, \ie Large gallery size, our approach outperforms SAVER by 5.5\% and 5.4\% on Top-1 and Top-5 accuracy, respectively.
\end{itemize}

\section{Self-supervised geometric features discovery via interpretable attention}
As illustrated in Figure \ref{fig-overview}, in order to learn self-supervised geometric features as well as global representations simultaneously, our framework is composed of \textit{Global Branch (GB)}, \textit{Self-supervised Learning Branch (SLB)} and \textit{Geometric Features Branch (GFB)}. Each branch has its own function and also interacts with each other. Generally, GB is employed to encode robust global codes from an input image. SLB performs the auxiliary self-supervised representation learning. By sharing its encoder with SLB, GFB is able to discover discriminative features from automatically discovered geometric locations without corresponding supervision. In remaining subsections, we elaborate each main component in turn.

\subsection{Problem setup}
Given a query image, vehicle ReID is to obtain a ranking list of all gallery images according to the similarity between query and each gallery image. The similarity score is typically calculated from deep embeddings, \ie{$cos(f(x_q;\theta),\;f(x_g;\theta))$}. Here $\;f(\cdot;\theta)$ represents a deep network with learnable parameters $\theta$; $x_q$, $x_g$ are query and gallery image respectively; $cos(\cdot)$ denotes cosine similarity computation. $\;f(\cdot;\theta)$ is optimized on a training set $D={\{x_i,\;y_i\}}_{i=1}^N$, where $x_i$, $y_i$ are a vehicle image and its identity label and $N$ is the number of training samples.

\subsection{Self-supervised learning for highlighting geometric features} \label{selfsupervised}
Self-supervised learning is equivalent to optimizing a deep network under the supervision of machine generated pseudo labels. Among them, image rotation degree prediction, \ie{rotating image by a random angle and training a classifier to predict it}, has demonstrated its capacity in many tasks \cite{gidaris2018unsupervised, Zhai_2019_ICCV, feng2019self, Kolesnikov_2019_CVPR}. Vehicle ReID can be regarded as an instance-level classification problem, \ie all images contain the same species but many instances. Thus salient object in each image has similar geometry properties, \eg shape, outline, and skeleton. We argue that training a network to predict the rotation degree of a randomly rotated vehicle image encourages it to focus on these reliable and shared geometric properties (it is the same for person ReID), which can help to easily recognize the rotation of an object. This geometric information has been proven crucial and discriminative for distinguishing a vehicle instance \cite{key_points, dual_path} although it was represented by manually annotated keypoints as shown in Figure \ref{fig-fig1} (a).  

Concretely, we first rotate an image $x_i$ from $D$ by $0^\circ$, $90^\circ$, $180^\circ$ or $270^\circ$ (assigning class 0, 1, 2 or 3 respectively) to generate a new dataset $D_{SL}={\{x_{i,r},\;y_r\}}_{r=1}^4,\;i=1,...,N$. Subsequently, the image $x_{i,r}$ is fed into an shared encoder $f_{ae}(\cdot;\theta_{ae})$ (namely attention encoder in Figure \ref{fig-overview}) to extract low-level semantics, $f_{ae}(x_{i,r};\theta_{ae})$. To predict rotation class, high-level representations need to be further condensed from $f_{ae}(x_{i,r};\theta_{ae})$. We append another deep module $f_{se}(\cdot;\theta_{se})$ to achieve this. Thus a high-dimensional embedding vector is obtained:
\begin{equation}
    F_{SL}(x_{i,r})=GAP\lbrack f_{se}(f_{ae}(x_{i,r};\theta_{ae});\theta_{se})\rbrack,
\end{equation} 
where $GAP\lbrack\cdot\rbrack$ denotes Global Average Pooling operation. To generate more compact clusters in embedded space, the Cosine Classifier (CC) \cite{boostfewshoot} is employed to assign the rotation class. The learnable parameters of CC is $W_{CC}=\left[w_1,\dots,w_j,\dots,w_b\right]$, $w_j\in\mathbb{R}^d$, where $d$ is the dimension of vector $F_{SL}$ and $b$ is the number of classes (\ie{$b=4$}). The probabilities of assigning the input image into each class can be represented as $P(x_{i,r})=\left[p_1,\dots,p_j,\dots,p_b\right]$, where each element is
\begin{equation}
    p_j=Softmax\left[\gamma\cos\left(F_{SL}(x_{i,r}),w_j\right)\right].
\end{equation}
$Softmax\left[\cdot\right]$ and $\gamma$ represent respective normalized exponential function and a learnable scalar. Finally, the objective function of self-supervised learning is: 
\begin{equation}
    {\mathcal L}_{SLB}={\mathbb{E}}_{D_{SL}}\lbrack CE(P(x_{i,r}),\;y_r)\rbrack,
\end{equation}
where $CE(\cdot)$ is Cross Entropy loss function. Obviously, the optimization of ${\mathcal L}_{SLB}$ enforces the deep classifier, especially the subnetwork $f_{ae}(\cdot;\theta_{ae})$, to capture geometric features from the input image.

\begin{figure}[t]
\centering
\includegraphics[width=0.99\columnwidth]{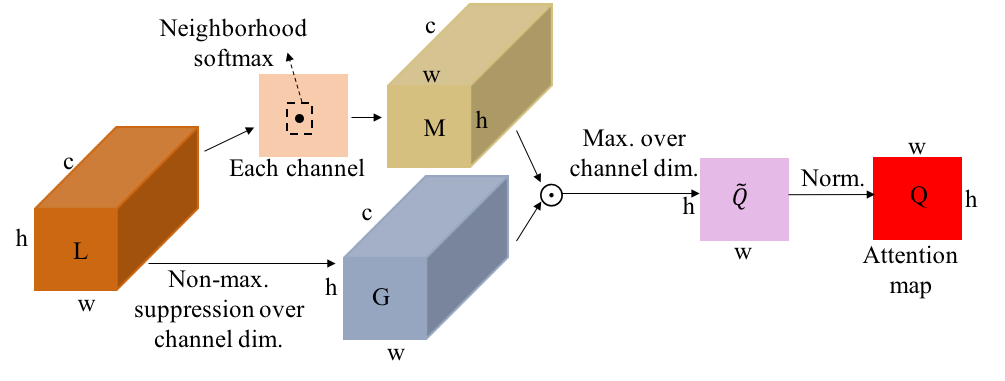} 
\vspace{-2mm}
\caption{Interpretable Attention Module (IAM).}
\vspace{-4.5mm}
\label{fig-acm}
\end{figure}

\subsection{Discriminative features discovery via interpretable attention}
Through performing self-supervised learning (Section \ref{selfsupervised}), low-level geometric features have been extracted by the shared encoder $f_{ae}(\cdot;\theta_{ae})$. We argue that the best way to discovering discriminative local patterns is to aggregate spatial locations of high response and concentrate on corresponding features of these points. Definitely, the former step is pretty important and a spatial attention may be a choice of achieving this. However, existing works in attention learning usually have two well-known drawbacks: (1) the unexplainable workflows, \ie the architectures were usually heuristically devised and their loads of parameters were completely learned by themselves;
(2) the scattered concerning areas, \ie the high response regions were too large to indicate discriminative patterns. Alternatively, we introduce an interpretable attention module whose deriving process can be reasonably explained and does not contain any learnable parameters. Furthermore, visualizations demonstrate that our attention can successfully focus on more accurate regions-of-interest that have physical meanings.

We illustrate the Interpretable Attention Module (IAM) in Figure \ref{fig-acm}, where $L\in\mathbb{R}^{c\times h\times w}$ is a 3D tensor extracted by $f_{ae}(\cdot;\theta_{ae})$ from an input image $x_i$ and $c, h, w$ denote the channel, height, and width dimension, respectively. To discover local points of interest on spatial dimensions, a $Softmax\left[\cdot\right]$ over neighborhood of each point is first conducted along each channel in $L$, \ie
\begin{align}
    M(k,u,v)=\frac{\exp\left(L(k,u,v)\right)}{\sum_{\left(m,n\right)\in\mathcal N(u,v)}\exp(L(k,m,n))},
\end{align}
where $\mathcal N(u,v)$ denotes the squared neighborhood set with side length $\mathcal{K}$ around location $(u,v)$ in the $k$-th channel. In parallel, a Non-Maximum Suppression (NMS) computing across all channels is performed from $L$ to highlight important feature channels, \ie
\begin{align}
G(k,u,v)=\frac{L(k,u,v)}{\max_{t=1,...,c}L(t,u,v)}.
\end{align}
To take the local spatial maxima and channel-wise maxima above into account altogether, $\Tilde{Q}$ is obtained by the element-wise product of $M$ and $G$ followed by the maximization over channel dimension, \ie $\Tilde{Q}(u,v)=\max_{t=1,...,c}\left\{M(t,u,v)\cdot G(t,u,v)\right\}$. Our final attention $Q$ is obtained by the spatial normalization of $\Tilde{Q}$, which considers all local maxima together and aggregates global points of interest: 
\begin{align}
    Q(u,v)=\frac{\Tilde{Q}(u,v)}{\sum_{(m,n)}\Tilde{Q}(m,n)}.
\end{align}
$Q$ represents the spatial emphasis of the activation tensor $L$, namely, crucial points of the input image $x_i$. So it is reasonable to weight another global representations extracted from $x_i$ with $Q$ as discriminative geometric features as in Figure \ref{fig-overview}. Our attention is partially inspired by the soft landmark detection in \cite{d2net} but significantly different from theirs. 

\subsection{Overall optimization objectives}
To optimize the whole framework, we combine CE loss from SLB, hard mining Triplet loss (Tri) \cite{triplet_loss} and Smoothed Cross Entropy loss (SCE) \cite{smoothed_ce1} from GFB and GB together as our final objectives. Tri and SCE loss are optimized referring to the combination mechanism of Batch Normalization Neck (BNNeck) \cite{bag_of_tricks}. Our overall objectives are 
\begin{align}\label{eqn:overall_loss}
    {\mathcal L}_{overall} = & \lambda_{Tri}^{GB}\mathcal L_{Tri}^{GB} + \lambda_{SCE}^{GB}\mathcal L_{SCE}^{GB} + \lambda_{Tri}^{GFB}\mathcal L_{Tri}^{GFB} \\ 
    & + \lambda_{SCE}^{GFB}\mathcal L_{SCE}^{GFB} + \lambda_{SLB}{\mathcal L}_{SLB} \nonumber.
\end{align}
To avoid heavy tuning of hyperparameters, we simply set the importance coefficients $\lambda_{Tri}^{GB}$, $\lambda_{SCE}^{GB}$, $\lambda_{Tri}^{GFB}$, $\lambda_{SCE}^{GFB}$ to 0.5 in all experiments. Only $\lambda_{SLB}$ is fine-tuned in ablation study and set to 1.0 in final experiments.

During inference, SLB is abandoned. Two feature vectors from GB and GFB are concatenated as representations of an input image.

\subsection{Network architecture}
We give the architecture configurations in Figure \ref{fig-overview} and each color represents a subnetwork. Referring to the literature, we choose ResNet50 \cite{resnet}, with $stride=2$ in $conv5\_x$ replaced with $stride=1$, as the backbone of GB. It is divided into two subnetworks, \ie the first ($conv1$, $conv2\_x$, $conv3\_x$) and the second ($conv4\_x$, $conv5\_x$), denoted by green and red respectively. The shared encoder between SLB and GFB is implemented by ResNet18 (orange) whose $stride$ in $conv4\_x$, $conv5\_x$ is set as 1. In SLB, another subnetwork (purple), consisting of two basic ResNet blocks \cite{resnet} with $stride=2$, is appended to the encoder to further condense features. In GFB, each image is first downsampled by 8 times by passing through the attention encoder and then the obtained tensor is processed by IAM to get the attention map. By element-wise multiplication, it is broadcast to every channel of the features from the first subnetwork of GB backbone, followed by another subnetwork (blue) composed of 
${conv4\_x}'$, ${conv5\_x}'$.

\section{Experiments} \label{sec:exp}
\bfsection{Datasets}
We conduct experiments on three vehicle ReID benchmarks. {\em VeRi-776} \cite{veri} contains 49,357 images of 776 vehicles and 37,778 images of 576 identities compose its training set. {\em CityFlow-ReID} \cite{cityflow} is a challenging dataset where images are captured by 40 cameras under diverse environments. 36,935 images from 333 identities form the training set. {\em VehicleID} \cite{vehicleid} is a large-scale benchmark containing 221,763 images of 26,267 vehicles. Its gallery set only contains one randomly selected image for each identity and thus we report our results as the mean over 10 trials. There are three numbers of gallery images widely used for testing, \ie{800 (Small), 1600 (Medium), and 2400 (Large)}.

\bfsection{Implementation}
We choose PyTorch to implement our framework and Adam optimizer \cite{adam} with default betas ($\beta_1=0.9$, $\beta_2=0.999$), weight decay 5e-4 to optimize it. During training, random cropping, horizontally flipping, and erasing are performed to augment data samples. None of them is adopted to process testing images. All images are resized to $256\times256$ and experiments are conducted on one NVIDIA GEFORCE RTX 2080Ti GPU. The batch size on VeRi-776 and CityFlow-ReID is 28 and that on VehicleID is 40, with 4 images from each instance. On VeRi-776 and CityFlow-ReID, the initial learning rate is 1e-4 and the margin of triplet loss is set as 0.5 empirically. The number of training epochs is 80 and the learning rate is decreased by a factor of 0.1 at 20th, 40th, and 60th epoch. On VehicleID, the margin is 0.7 and the number of learning epochs is 120. The learning rate is increased linearly from 0 to 1e-4 during the first 10 epochs, decreased with cosine scheduler to 1e-7 at 100th epoch, and to 0 at the last epoch. 
 
\bfsection{Evaluation protocols} Unlike some previous methods, we do not use any post-processing techniques like $k$-reciprocal re-ranking \cite{re_ranking} to refine our results. We evaluate our approach by four widely used metrics in ReID literature, \ie{image-to-track retrieval mean Average Precision (tmAP) (if tracks are available in one dataset), image-to-image retrieval mAP (imAP), Top-1, and Top-5 accuracy}. Particularly, we report both tmAP and imAP on VeRi-776 for comprehensive evaluation. These scores are shown as percentages and the best are marked in bold. In Table \ref{veri_table}, \ref{cityflow_table}, and \ref{vehicleid_table}, ES (Y/N) indicates whether Extra Supervision besides ID labels is employed to train a corresponding method.

\begin{table}[t]
\centering
\resizebox{0.99\columnwidth}{!}{%
\begin{tabular}{c|c|c|c|c|c|c} 
\toprule
Method     & Venue  & ES & tmAP & imAP & Top-1 & Top-5  \\ 
\hline\hline
OIFE \cite{key_points}   & ICCV17     & Y           & 48.0  & -       & 65.9     & 87.7         \\
OIFE+ST \cite{key_points}  & ICCV17   & Y           & 51.42  & -       & 68.3     & 89.7         \\
NuFACT \cite{liu2017provid}  & TMM17   & Y           & 53.42  & -       & 81.56     & 95.11      \\
VAMI \cite{Zhou_2018_CVPR}   & CVPR18     & Y           & 50.13  & -       & 77.03     & 90.82         \\
AAVER \cite{dual_path}  & ICCV19    & Y           & 58.52   & -      & 88.68     & 94.10      \\
RS \cite{pose_multi}   & ICCV19   & Y           & -  & 63.76       & 90.70     & 94.40      \\
R+MT+K \cite{pose_multi}  & ICCV19    & Y           & -  & 65.44       & 90.94     & 96.72      \\
VANet \cite{vanet}  & ICCV19    & Y           & 66.34  & -       & 89.78     & 95.99      \\
PART \cite{part_regularized}    & CVPR19    & Y          & 74.3  & -       & 94.3     & \textbf{98.7}      \\
SAN \cite{san}    & MST20     & Y           & 72.5   & -      & 93.3     & 97.1      \\
CFVMNet \cite{CFVMNet}    & MM20     & Y           & -   & 77.06      & 95.3     & 98.4      \\
PVEN \cite{parsing2020}   & CVPR20    & Y          & -  & 79.5       & 95.6     & 98.4      \\
SPAN \cite{chen2020orientation}  & ECCV20     & Y          & 68.9  & -       & 94.0     & 97.6      \\
\hline
DMML \cite{deep_meta}  & ICCV19   & N           & -  & 70.1       & 91.2     & 96.3      \\
UMTS \cite{umts}  & AAAI20   & N           & -  & 75.9       & 95.8     & -      \\
SAVER \cite{khorramshahi2020devil}  & ECCV20   & N           & 79.6  & -       & 96.4     & 98.6      \\  
\textbf{Ours}  & -   & N            & \textbf{86.2}     & \textbf{81.0}          & \textbf{96.7}            & 98.6                \\
\bottomrule
\end{tabular}
}
\caption{Results comparison on VeRi-776.}
\vspace{-3mm}
\label{veri_table}
\end{table}

\subsection{Performance comparison with SOTA works}
\bfsection{VeRi-776} 
We compare our approach with SOTA ones in Table \ref{veri_table}. We can see most works utilized extra supervisions to achieve their performances. For instance, VANet annotated 5,000 images from each dataset to train a viewpoint predictor and learned distinct metrics for similar and dissimilar viewpoint pairs. PART defined three types of vehicle parts, \ie lights, windows, and brands, to train a YOLO \cite{redmon2016you}. When training the ReID model, they extracted local features from detected regions by YOLO as supplementary information for global representations. PVEN provided view segmentations of 3,165 images to train a U-Net segmentor, whose output mask was used for view-aware feature alignment when optimizing their model. Although no enhancement from extra labels was utilized in SAVER, Detectron \cite{girshick2018detectron} was needed to pre-process all images to remove background noise. In contrast, our approach does not involve any extra annotations to assist local feature learning. Although our training batch size 28 is much smaller than other methods (\eg 256 in SAN), our method can still outperform other competitors significantly on tmAP and imAP. Regarding Top-5 accuracy, ours is only 0.1\% lower than the best that used a larger image size $512\times512$. That was demonstrated to promote their performances considerably \cite{part_regularized}. When comparing under the same condition, ours are much better than theirs on all indicators. 

\begin{table}[t]
\centering
\resizebox{0.99\columnwidth}{!}{%
\begin{tabular}{c|c|c|c|c|c} 
\toprule
Method    & Venue & ES & imAP & Top-1 & Top-5  \\ 
\hline\hline
FVS \cite{fvs}   & CVPRW18    & Y             & 5.08      & 20.82     & 24.52      \\
RS \cite{pose_multi}    & ICCV19   & Y           & 25.66         & 50.37     & 61.48      \\
R+MT+K \cite{pose_multi}  & ICCV19     & Y           & 30.57         & 54.56     & 66.54      \\
SPAN \cite{chen2020orientation}  & ECCV20     & Y             & \textbf{42.0}      & 59.5     & 61.9      \\  
\hline
Xent \cite{torchreid} & arXiv19 & N             & 18.62        & 39.92     & 52.66      \\
Htri \cite{torchreid}  & arXiv19       & N             & 24.04       & 45.75     & 61.24         \\
Cent \cite{torchreid}  & arXiv19    & N             & 9.49       & 27.92     & 39.77         \\
Xent+Htri \cite{torchreid}  & arXiv19       & N             & 25.06       &51.69     & 62.84         \\
BA \cite{BABS}  & IJCNN19     & N             & 25.61       & 49.62     & 65.02     \\
BS \cite{BABS}  & IJCNN19     & N             & 25.57      & 49.05     & 63.12      \\
\textbf{Ours}  & - & N                 & 37.14          & \textbf{60.08}            & \textbf{67.21}                \\
\bottomrule
\end{tabular}
}
\caption{Results comparison on CityFlow-ReID.
}
\vspace{-4mm}
\label{cityflow_table}
\end{table}

\bfsection{CityFlow-ReID} 
The results are reported in Table \ref{cityflow_table}. This dataset is quite challenging because images are taken from five scenarios, covering a diverse set of location types and traffic conditions. Results of metric learning methods (Xent, Htri, Cent, Xent+Htri) and batch-based sampling ones (BA, BS) are acquired without using extra annotations. To assist ReID, real and synthetic images were exploited by RS, while R+MT+K employed keypoints, vehicle type, and color class to perform multi-task learning. SPAN adopted vehicle orientation information to guide visible feature extraction and computed a co-occurrence part-attentive distance for each image pair. As we see, except for SPAN, our approach surpasses others by large margins on all three metrics, \eg $\sim$7.0\% imAP, $\sim$6.0\% Top-1, and $\sim$1.0\% Top-5 accuracy compared with R+MT+K.

\setlength{\tabcolsep}{2pt}
\begin{table}[t]
\centering
\resizebox{0.99\columnwidth}{!}{%
\begin{tabular}{c|c|c|c|c|c|c|c|c} 
\toprule
\multirow{2}{*}{Method} & \multirow{2}{*}{Venue} & \multirow{2}{*}{ES} & \multicolumn{2}{c|}{Small} & \multicolumn{2}{c|}{Medium} & \multicolumn{2}{c}{Large}  \\ 
\cline{4-9}
                &        &         & Top-1 & Top-5      & Top-1 & Top-5       & Top-1 & Top-5      \\ 
\hline\hline
GoogLeNet \cite{cars_cuhk}   & CVPR15         & Y                            & 47.90      & 67.43               & 43.45          & 63.53                & 38.24      & 59.51           \\
MD+CCL \cite{vehicleid} & CVPR16     & Y                            & 49.0      & 73.5               & 42.8          & 66.8                & 38.2      & 61.6           \\
OIFE \cite{key_points}  & ICCV17    & Y                            & -       &-                & -          & -                & 67.0      & 82.9           \\
NuFACT \cite{liu2017provid}    & TMM17      & Y                            & 48.90      & 69.51               & 43.64          & 65.34                & 38.63      & 60.72           \\
VAMI \cite{Zhou_2018_CVPR}   & CVPR18       & Y                            & 63.1      & 83.3               & 52.9          & 75.1                & 47.3      & 70.3           \\
AAVER \cite{dual_path}  & ICCV19         & Y                            & 72.47      & 93.22               & 66.85          & 89.39                & 60.23      & 84.85              \\
VANet \cite{vanet}     & ICCV19         & Y                            & \textbf{88.12}      & 97.29              & 83.17          & 95.14                & 80.35     & 92.97              \\
PART \cite{part_regularized}   & CVPR19        & Y                            & 78.4      & 92.3               & 75.0          & 88.3                & 74.2      & 86.4           \\
SAN \cite{san}      & MST20             & Y                            & 79.7      & 94.3               & 78.4          & 91.3                & 75.6      & 88.3              \\
CFVMNet \cite{CFVMNet}      & MM20             & Y                            & 81.4      & 94.1               & 77.3          & 90.4                & 74.7      & 88.7              \\
PVEN \cite{parsing2020}   & CVPR20        & Y                            & 84.7      & 97.0               & 80.6          & 94.5                & 77.8      & 92.0           \\
\hline
UMTS \cite{umts}   & AAAI20       & N          &80.9 &- &78.8 &- &76.1 &-           \\
SAVER \cite{khorramshahi2020devil}   & ECCV20       & N          &79.9 &95.2 &77.6 &91.1 &75.3 &88.3           \\      
\textbf{Ours}  & - & N          & 86.8            & \textbf{97.4}          & \textbf{83.5}          & \textbf{95.6}          & \textbf{80.8}            & \textbf{93.7}          \\

\bottomrule
\end{tabular}
}
\caption{Results comparison on VehicleID.}
\label{vehicleid_table}
\end{table}

\bfsection{VehicleID}  
We list the results for comparison in Table \ref{vehicleid_table}. Note that VANet and PVEN required much larger batch size 128 and 256, respectively. Even so, our approach beats all competitors in almost every test setting. In particular, compared with SAVER, ours achieves much better performances on all gallery sizes, \ie{6.9\% Top-1, 2.2\% Top-5 on Small, 5.9\% Top-1, 4.5\% Top-5 on Medium, 5.5\% Top-1, 5.4\% Top-5 on Large higher}, although it involved some specific pre-processing steps.

\begin{figure}[t]
\centering
\includegraphics[width=0.99\columnwidth]{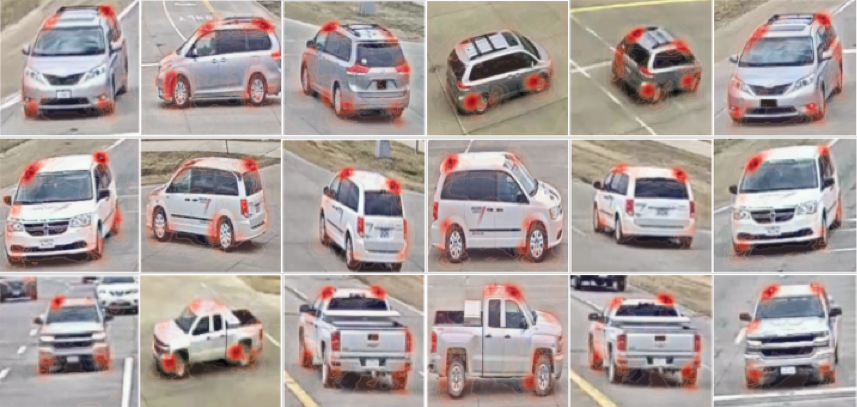} 
\caption{Discovered consistent geometric features from various viewpoints of the same vehicle (each row).
}
\vspace{-3mm}
\label{fig-consistency}
\end{figure}

\subsection{Visualizations of discovered geometric features through self-supervision}
We cover an input image with its attention map from GFB to visualize critical vehicle parts learned by our framework. Even though our geometric features are discovered without using accurate supervision like others, qualitative visualizations demonstrate the superiority of our method.

\bfsection{Comparison with defined landmarks by other works}
Previous works manually annotated a specific number of landmarks on a vehicle body \cite{dual_path, key_points} to assist their discriminative feature learning. These landmarks visible from each viewpoint (front, back, left or right) are illustrated in Figure \ref{fig-fig1} (a) which is borrowed from \cite{key_points} and re-organized vertically. To compare with these human annotations thoroughly, we visualize our learned geometric features from each viewpoint accordingly in Figure \ref{fig-fig1} (b). We can easily observe that our approach focuses on many similar locations to predefined ground truths on each viewpoint, \eg \textit{left (right)-front corner of vehicle top, left (right) fog lamp} on front view, \textit{right-front corner of vehicle top, right-front (back) wheel, and right headlight} on right view, which demonstrates that our framework can successfully discover critical and informative vehicle parts for ReID without the supervision of ground truths. 

\bfsection{Consistency across viewpoints and scenarios}
To validate the consistency of learned geometric features across viewpoints and scenarios, we select a couple of images, belonging to an identical vehicle instance but taken from various viewpoints and by different cameras, for visualization in Figure \ref{fig-consistency}. Each row represents an vehicle instance. Although viewpoint, object scale, and background of each image vary largely, identical vehicle parts, \eg{fog lamps, vehicle tops, and wheels}, are discovered for the same instance. This validates the stability and reliability of our approach in handling viewpoint and scenario changes, which are the key points of solving ReID problems. 

\bfsection{Generalization to human part discovery}
To demonstrate the generalization ability of our framework to person ReID, we conduct experiments on two popular benchmarks. Please refer to Section \ref{generalization} for more experiment details and here we just analyze the visualization results shown in Figure \ref{fig-fig1} Bottom. As a deformable object, discovering geometric features from a human body is much more challenging. For saving space, we just select three images for each person. Obviously, identical human parts, \eg head, upper arms, and knees, are discovered by our approach even though human pose, viewpoint, and background change so much among images. These parts are critical to estimate human pose which has been demonstrated to play an important role in person ReID \cite{Xu_2018_CVPR, Liu_2018_CVPR, su2017pose}.

\bfsection{Discussion}
As mentioned in Section \ref{selfsupervised}, person or vehicle ReID is an instance classification problem, \ie all images in a task are taken from the same category but different individuals. So salient objects in these images have a lot in common, \eg geometric shapes (for vehicles), compositions, and skeletons. It is reasonable that conducting self-supervised learning encourages a deep network to discover these geometric features because they are reliable and repeatable clues to completing the self-supervised pretext task successfully. Visualizations in this section demonstrate this claim sufficiently. In view of their high similarity to ReID, we will expand our approach to other fine-grained classification tasks \cite{wah2011caltech, maji2013fine} in future work. 

\begin{table}[t]
\centering
\resizebox{0.9\columnwidth}{!}{%
\begin{tabular}{c|c|c|c|c}
\toprule
\multirow{2}{*}{Method}                                                          & \multicolumn{2}{c|}{VeRi-776} & \multicolumn{2}{c}{CityFlow-ReID} \\ 
\cline{2-5}
                                                                                 & tmAP & imAP   & imAP & Top-1\\ 
\hline\hline
GB w/o attention                                                                                & 84.0            & 78.3              & 32.04            & 56.27                \\
GB+ResNet18 w/o attention                                                                               & 85.2            & 79.5              & 34.63            & 57.98 \\
\hline
GB+GFB ($\mathcal K=7$)  & \underline{85.9}            & \underline{80.7}              & \underline{36.63}            & \underline{59.98}       \\
GB+GFB ($\mathcal K=11$) & 85.8            & 80.6              & 35.94            & 59.70       \\
GB+GFB ($\mathcal K=15$) & 85.5            & 80.2              & 36.32            & 58.56  \\ 
\hline
GB+GFB+SLB ($\lambda_{SLB}=0.1$)                         & 85.8            & 80.5              & 36.61            & 59.13                \\
GB+GFB+SLB ($\lambda_{SLB}=1.0$)                         & \underline{\textbf{86.2}}            & \underline{\textbf{81.0}}              & \underline{\textbf{37.14}}            & \underline{\textbf{60.08}}  \\
GB+GFB+SLB ($\lambda_{SLB}=2.0$)                         & 86.1            & 80.9              & 36.54            & 59.60                \\
\bottomrule
\end{tabular}
}
\caption{Results of ablation study. We underline the corresponding results of selected values for $\mathcal K$ and $\lambda_{SLB}$. The performance improvement upon each component is consistent across datasets.}
\label{ablation_table}
\end{table}

\begin{figure}[t]
\centering
\includegraphics[width=0.99\columnwidth]{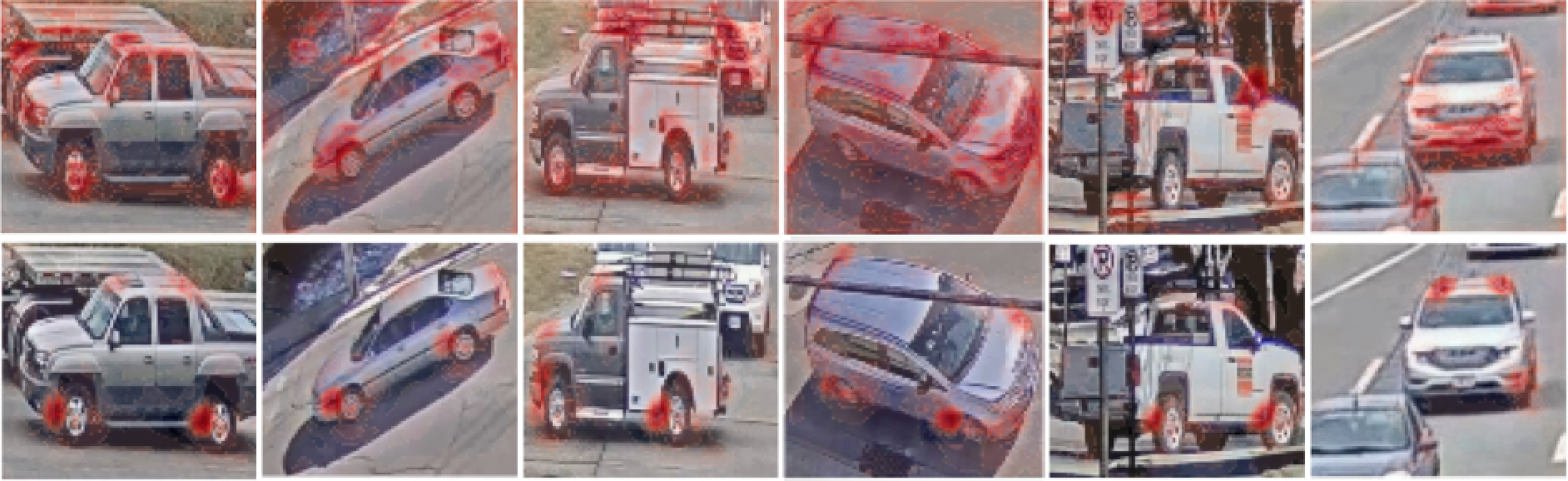}
\vspace{-2mm}
\caption{Learned attention maps (top) without and (bottom) with self-supervised learning from the same image.
}
\vspace{-4mm}
\label{fig:vis_comp}
\end{figure}

\subsection{Ablation study}
To evaluate the effect of each proposal of our framework, we conduct extensive experiments on VeRi-776 and CityFlow-ReID. Here we report tmAP and imAP, imAP and Top-1 on them respectively because these metrics are more important on each dataset. Results are in Table \ref{ablation_table}. 

\bfsection{Effect of simply incorporating another branch upon baseline}
Our framework employs a ResNet50 as the backbone of GB and a ResNet18 as the shared encoder between SLB and GFB. Although more branches and larger networks were usually utilized to perform ReID in previous works, we still conduct experiments to show that our performance gains upon the baseline (GB w/o attention) come from our proposals rather than an extra branch. To this end, we implement a new framework termed as ``GB+ResNet18 w/o attention'',  consisting of two independent branches with respective ResNet50 and ResNet18 as backbone. Compared with our final results, we can see performances are marginally beneficial from adding a ResNet18 based branch. However, this also suggests that our framework can be stronger if involving more branches like others.

\bfsection{Pure IAM still bringing much improvement}
IAM based GFB is the bridge of incorporating self-supervised learning (SLB) into our whole framework. To demonstrate the effectiveness of IAM even without the regularization from SLB, we conduct experiments ``GB+GFB'' with different $\mathcal K$ while fixing other hyperparameters. Results in the second part of Table \ref{ablation_table} show that IAM are pretty robust \wrt values of $\mathcal{K}$. We set $\mathcal K=7$ by default in subsequent experiments. Besides, comparing ``GB+GFB'' with ``GB+ResNet18 w/o attention'', it is seen that our IAM based GFB is much more powerful than a ResNet18 branch. For example, on the challenging CityFlow-ReID, the former and the latter bring improvement about 4.6\% \vs{2.6\%} on imAP and 3.7\% \vs{1.7\%} on Top-1 accuracy upon the baseline. This is attributed to the interpretable attention IAM. Although it is not able to discover specific parts without the help of self-supervised learning (referring to Figure \ref{fig:vis_comp}), it can focus on a holistic vehicle body from cluttered background for extracting more effective representations.

\bfsection{Physically meaningful attention discovery through self-supervised learning}
To enforce attention to emphasize crucial vehicle parts (\ie physically meaningful locations), we conduct experiments with the full framework ``GB+GFB+SLB'' with different $\lambda_{SLB}$ while keeping other hyperparameters identical. The third part results in Table \ref{ablation_table} tell that our framework are robust \wrt values of $\lambda_{SLB}$ and we select $\lambda_{SLB}=1.0$ as our decision. It is observed that self-supervised learning improves performances consistently on all metrics compared with ``GB+GFB''. Especially, from the attention maps comparison in Figure \ref{fig:vis_comp}, we can see self-supervised learning helps to shift attentions distracted by background vehicles to the main concern. And our framework overcomes the interference from diverse background distractors, \eg traffic lights and road signs, and discovers meaningful vehicle parts successfully.

\subsection{Generalizing to other ReID related tasks} \label{generalization}
In this section, we demonstrate the potential application of our approach in person ReID and multi-target multi-camera (MTMC) vehicle tracking.

\bfsection{Person ReID}
Instead of identifying individual vehicles, this task aims to associate the same person in images taken from different cameras. We conduct experiments on Market-1501 \cite{market1501} and DukeMTMC-reID \cite{dukemtmcreid}, two most widely used benchmarks for person ReID. The training details keep identical to those on VehicleID. We compare our performances with recent works in Table \ref{person_table}. As we see, though our approach is not intentionally proposed and tuned for person ReID, its performances are still very promising. We believe it will perform much better if the hyperparameters are fine-tuned accordingly.

\begin{table}[t]
\centering
\resizebox{.9\columnwidth}{!}{%
\begin{tabular}{c|c|c|c|c|c|c|c} 
\toprule
\multirow{2}{*}{Method} & \multirow{2}{*}{Venue} & \multicolumn{3}{c|}{Market-1501} & \multicolumn{3}{c}{DukeMTMC-reID}  \\ 
\cline{3-8}
 & & imAP & Top1 & Top5     & imAP & Top1 & Top5          \\ 
\hline\hline
DG-Net \cite{dgnet}      & CVPR19          & 86.0               & 94.8   & -       & 74.8                & 86.6     & -      \\
Bag \cite{bag_of_tricks}   & TMM19        & 85.9               & 94.5   & -       & \textbf{76.4}                & 86.4     & -      \\
PCB \cite{pcb}      & ECCV18       & 81.6               & 93.8   & 97.5       & 69.2                & 83.3     & 90.5      \\
RGA \cite{rga}    & CVPR20           & \textbf{88.4}               & \textbf{96.1}   & -       & -                & -     & -      \\
OSNet \cite{osnet}     & ICCV19        & 84.9               & 94.8   & -       & 73.5  & \textbf{88.6}     & -      \\
\hline

\textbf{Ours} & -  & 86.1          & 94.3  & \textbf{98.3}        & 75.7            & 85.7    & \textbf{93.6}      \\

\bottomrule
\end{tabular}}
\caption{Results comparison on person ReID benchmarks.
}
\label{person_table}
\vspace{-2mm}
\end{table}

\bfsection{MTMC vehicle tracking}
As a complicated video surveillance task, MTMC vehicle tracking is commonly composed of four steps, \ie{vehicle detection, multi-target single-camera (MTSC) tracking, vehicle re-identification, and tracklet synchronization.} Among them, vehicle ReID is the crucial stage for a satisfactory tracking result. It is much more challenging than operating on well-calibrated ReID benchmarks because of large object-scale variation and heavy blur caused by distance changes between a camera and vehicles. To verify the generalization ability of our approach under cross-dataset testing, we perform experiments on the data provided by City-Scale Multi-Camera Vehicle Tracking of AI City 2020 Challenge \cite{aicitychallenge} using our trained model on VeRi-776 without any fine-tuning. Considering that ReID is only our concern, we simply adopt an efficient MTMC tracking pipeline, similar to \cite{electricity}, to achieve this. Specifically, we first employ Mask R-CNN from Detectron2 \cite{wu2019detectron2} to detect vehicles from each video frame. Then we utilize Deep SORT \cite{deepsort} with the association strategy from \cite{wang2019towards} to perform MTSC vehicle tracking. Finally, our trained model is directly applied to capture ReID representations from cropped vehicle images, followed by tracklet synchronization with identical rules to \cite{electricity}. Refer to \cite{electricity} for more details due to page limitation. Our approach achieves 0.4930 regarding the official evaluation metric IDF1 score \cite{cityflow}, which is much higher than 0.4585 from \cite{electricity}, although they trained their ReID model on the officially provided dataset. Besides, we compare our result with other submissions in Table \ref{leaderboard} and ours outperforms others significantly.

\begin{table}[t]
\centering
\resizebox{0.99\columnwidth}{!}{
\begin{tabular}{c|c|c|c|c|c|c} 
\toprule
Rank & 1 & 2 & 3 & 4 & 5 & 6                                \\ 
\hhline{=======}
Team ID  & Ours & 92 & 141 & 11 & 163 & 63\\
\hline
IDF1 Score    & \textbf{0.4930}  & 0.4616 & 0.4552 & 0.4400 & 0.4369 & 0.3677\\
\bottomrule
\end{tabular}
}
\caption{MTMC vehicle tracking results comparison on AI City Challenge 2020.}
\label{leaderboard}
\vspace{-5mm}
\end{table}

\section{Conclusion}
In this paper, based on our observation that salient objects in ReID images share similar properties, we propose a novel framework to learn geometric features, without supervision from fine-grained annotations, for vehicle ReID through performing a self-supervised task. To this end, an interpretable attention module is also introduced to discover physically reasonable features. Comprehensive experiments demonstrate the effectiveness and generalization ability of our approach qualitatively and quantitatively. In the future, we plan to generalize it to addressing fine-grained classification problems.

\newpage
{\small

\bibliographystyle{ieee_fullname}
\bibliography{egbib}
}

\end{document}